# Transformer in Touch: A Survey

Jing Gao[1], Ning Cheng[1], Bin Fang[2], and Wenjuan Han[1,3,4,✉]

[1] Beijing Key Lab of Traffic Data Analysis and Mining, Beijing Jiaotong University, Beijing. China
[2] School of Artificial Intelligence, Beiing University of Posts and Telecommunications, Beijing, China
[3] China Railway Design Corporation, Tianjin, China
[4] National Engineering Research Center for Digital Construction and Evaluation of Urban RailTransit,Tianjin. China
wjhan@bjtu.edu.cn

**Abstract.** The Transformer model, initially achieving significant success in the field of natural language processing, has recently shown great potential in the application of tactile perception. This review aims to comprehensively outline the application and development of Transformers in tactile technology. We first introduce the two fundamental concepts behind the success of the Transformer: the self-attention mechanism and large-scale pre-training. Then, we delve into the application of Transformers in various tactile tasks, including but not limited to object recognition, cross-modal generation, and object manipulation, offering a concise summary of the core methodologies, performance benchmarks, and design highlights. Finally, we suggest potential areas for further research and future work, aiming to generate more interest within the community, tackle existing challenges, and encourage the use of Transformer models in the tactile field.

## 1   Introduction

In recent years, the Transformer model [1] has caused a sensation in the world of deep learning. In particular, the Transformer model has revolutionized diverse fields with its attention mechanism. This mechanism allows the model to focus on different parts of the input sequence, capturing long-range dependencies and improving performance on various tasks. The Transformer's ability to handle sequential data has paved the way for significant advancements, especially in the fields of natural language processing (NLP) and computer vision (CV) [2].

In the field of NLP, Transformer continues to be a major driving force for technological advancement [3, 4]. Apart from the well-known BERT [5] and GPT-3 [6], newer Transformer-based models have shown even stronger performance. For instance, GPT-4 [7] offers more refined language understanding and generation capabilities, applied to more complex language understanding, text generation, fine-grained sentiment analysis, and other tasks; other models such as ELECTRA [8] and DeBERTa [9] have also shown outstanding performance, achieving unprecedented results in multiple NLP benchmarks. These models optimize the training process and parameter efficiency, achieving better performance with fewer parameters, demonstrating the ongoing innovation and potential of Transformer models in the NLP field. Additionally, cross-language models, like mT5 [10] and XLM-R [11], utilizing Transformer technology, have greatly advanced research in machine translation and cross-cultural communication. All the aforementioned examples demonstrate the enormous potential of Transformer models in handling complex, large-scale datasets, and their key role in the future development of NLP.

Similar to NLP, in the field of CV, Transformer models have demonstrated their unique ability, particularly in processing spatial and temporal continuity in image data. This technology has been widely applied to various tasks such as image recognition [12], object detection [13], image segmentation [14], image super-resolution [15], video understanding [16], image generation [17], text-image synthesis [18], and visual question answering [19]. For example, models using Transformer technology can create unprecedentedly realistic and innovative images in image generation tasks [20], and in video understanding, they can identify and interpret dynamic images more accurately [21].

However, in the field of tactile technology, the application of Transformer models is still in its early stages. Tactile technology plays a crucial role in understanding the state and characteristics of objects, such as their shape, posture, texture, and material [22, 23, 24, 25, 26, 27]. Optical tactile sensors like GelSight [28], GelTip [29], TacTip [30], and DIGIT [31] offer higher-resolution tactile images compared to traditional single-point touch sensors and tactile arrays. These sensors are extensively used in various robot tasks, including sliding detection [32], texture recognition [33], surface tracking [34], object pushing [35], cable manipulation [36], insertion and tightening [37], and more.

Through these sensors, previous studies, such as [34, 35], often used tactile textures to extract handcrafted features, but these features were less effective when dealing with diversified and complex textured objects.

Recent research has taken a different approach by exploring end-to-end methods for learning tactile-motor strategies [32, 36, 37]. These methods directly extract features from tactile images using convolutional neural networks (CNNs), recurrent neural networks (RNNs), or Transformers, which then guide robot manipulation strategies. By enhancing the gradient consistency of the neural network, these methods improve the ability of tactile-motor strategies to generalize across different objects. Among them, approaches based on Transformer architectures have demonstrated significant promise. They excel in achieving robust and efficient tactile perception and manipulation within the field of robotics [38, 39].

In light of the advantages of Transformers in the tactile field, this review aims to comprehensively introduce the research progress in the tactile field adopting Transformer architecture, especially in applications with two types of inputs: Visuo-Tactile (a.k.a., vision-touch) and Only-Tactile (i.e., touch only). We will delve into various tasks in the tactile field, such as object recognition, cross-modal generation, object manipulation, etc. (Sec. 4), and the effects of using Transformer-based models. Through this article, complemented by the structured overview provided in the mind map (Fig. 1), we aim to offer a detailed reference resource for researchers in the tactile field to understand and utilize Transformer architecture and to provide guidance for future research directions. Additionally, the article will discuss the challenges faced by Transformer architecture in the tactile field and the direction of future developments, as outlined in the thematic areas of our mind map.

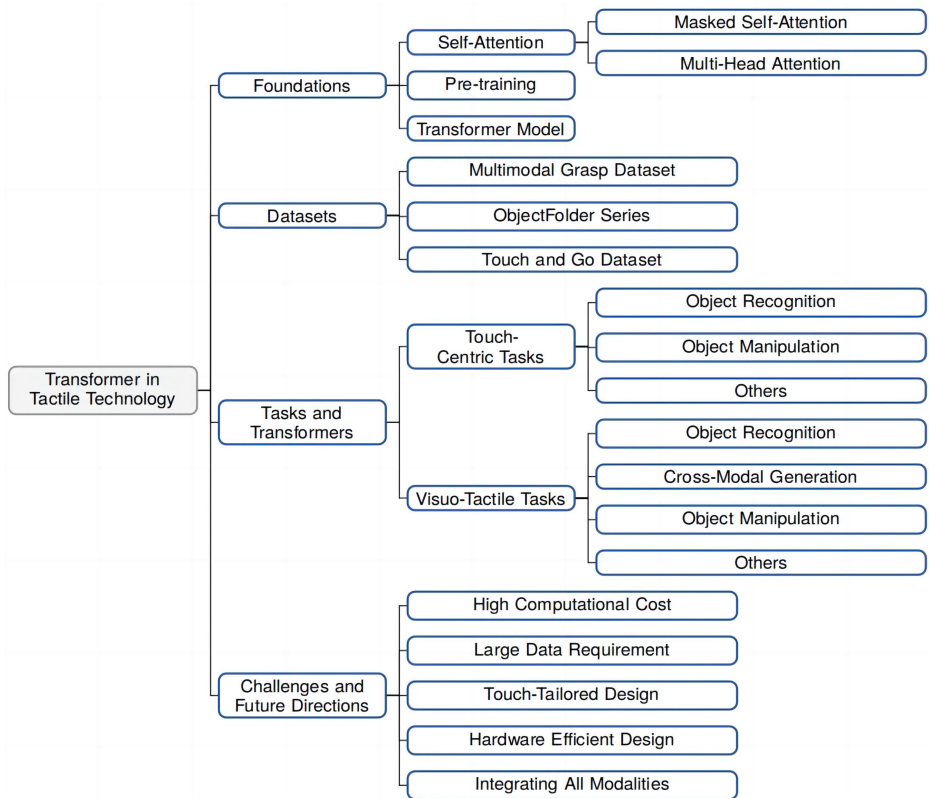

**Fig. 1.** Typology of Transformer in Tactile Technology

## 2 Foundations

In the context of developing Transformer models, two fundamental ideas significantly contribute to their evolution and progress [40]. First, the self-attention mechanism allows for capturing long-term dependencies between sequence elements, which is difficult for traditional recurrent models to achieve. Compared to recurrent models, the self-attention mechanism can encode these relationships more effectively. Second, pretraining is another core idea, referring to training on a large-scale labeled or unlabeled corpus in a self-supervised or supervised manner, followed by fine-tuning on relatively small labeled datasets for target tasks [5, 41, 42]. These two ideas have played a vital role in the development of the Transformer model and have been widely applied in various Transformer networks.

Next, we will provide a brief introduction to these two key ideas (Sec. 2.1 and Sec. 2.2, respectively), and summarize Transformer networks that apply these ideas (Sec. 2.3). This prior knowledge will assist us in gaining a better understanding of the Transformer-based models utilized in the tactile field (Sec. 4).

## 2.1    Self-Attention in Transformers

For an item in a sequence, self-attention estimates the relevance of the item with other items (for example, which words in a sentence are likely to occur together). Self-attention is a core component of the Transformer, which is pivotal for modeling the interactions between all elements in a sequence. Essentially, a self-attention layer updates each item of the sequence by aggregating global information from the entire input sequence. Suppose a sequence contains $n$ entities, represented as $(x_1, x_2, ..., x_n)$, where $d$ is the embedding dimension used to represent each entity. The goal of self-attention is to capture the interactions between all $n$ entities by encoding global contextual information for each entity. This is achieved by defining three learnable weight matrices for transforming Queries ($W^Q \in \mathbb{R}^{d \times d_q}$), Keys ($W^K \in \mathbb{R}^{d \times d_k}$) and Values ($W^V \in \mathbb{R}^{d \times d_v}$). The input sequence $X$ is first projected onto these weight matrices, obtaining $Q = XW^Q$, $K = XW^K$, and $V = XW^v$. The output $Z \in \mathbb{R}^{n \times d_v}$ of the self-attention layer is

$$\mathbf{Z} = \text{softmax}\left(\frac{\mathbf{QK}^T}{\sqrt{d_q}}\right)\mathbf{V}. \tag{1}$$

For a given entity in the sequence, self-attention first calculates the dot product of the query with all keys, followed by normalization with the softmax operator to obtain attention scores. Then, each entity becomes the weighted sum of all entities in the sequence, where weights are given by the attention scores.

**Masked Self-Attention.** The standard self-attention layer focuses on all entities. For the Transformer model [1], which is trained to predict the next entity in the sequence, the self-attention block used in the decoder is masked to prevent attention to subsequent future entities. This is simply accomplished by an element-wise multiplication operation with a mask $\mathbf{M} \in R^{n \times n}$, where $M$ is an upper triangular matrix. Masked self-attention is defined as

$$\text{softmax}\left(\frac{\mathbf{QK}^T}{\sqrt{d_q}} \circ \mathbf{M}\right) \tag{2}$$

where ∘ represents the Hadamard product. Essentially, when predicting an entity in the sequence, masked self-attention sets the attention scores for future entities to zero.

**Multi-Head Attention.** To encapsulate the multiple complex relationships between different items in a sequence, multi-head attention contains multiple self-attention blocks (in the original Transformer model [1], the number of self-attention blocks $h = 8$). Each block has its own set of learnable weight matrices $\{\mathbf{W}^{Q_i}, \mathbf{W}^{K_i}, \mathbf{W}^{V_i}\}$, where $i = 0 \cdots (h-1)$. For the input $X$, the outputs of the $h$ self-attention blocks in multi-head attention are then concatenated into a single matrix $[\mathbf{Z_0}, \mathbf{Z_1}, \cdots \mathbf{Z_{h-1}}] \in \mathbb{R}^{n \times h \cdot d_v}$ and projected onto a weight matrix $\mathbf{W} \in \mathbb{R}^{h \cdot d_v \times d}$.

The main difference between self-attention and convolution operations is that the filters are dynamically computed, rather than static filters in convolutions (which remain unchanged for any input). Additionally, self-attention is invariant to permutations and changes in the number of input points. Therefore, it can easily handle irregular inputs, while standard convolutions require a grid structure.

## 2.2 (Self) Supervised Pre-training

Transformer models based on self-attention typically employ a two-stage training mechanism. First, they undergo pre-training on large-scale datasets, which can be supervised or self-supervised. Subsequently, the pre-trained weights are adapted to small or medium-sized downstream task datasets.

Examples of downstream tasks in natural language processing include question answering, while in computer vision, tasks like image classification and object detection are common. In the fields of NLP and CV, the effectiveness of large-scale Transformer pre-training has been demonstrated. For instance, the Vision Transformer model (ViT-L) [12] saw a 13% absolute drop in accuracy on the ImageNet test set when trained only on the ImageNet training set (which has 50,000 images), compared to pre-training on the JFT dataset (which has 300 million images).

Since obtaining manual labels on a large scale is cumbersome, self-supervised learning has been very effectively used in the pre-training phase. Self-supervised pre-training plays a key role in enhancing the scalability and generalization abilities of Transformer networks, enabling them to train networks with over a trillion parameters (e.g., Google's Switch Transformer [43]).

Self-supervised learning offers a promising learning paradigm as it enables learning from large amounts of readily available unlabeled data. In the self-supervised pre-training phase, the model is trained to learn meaningful representations of underlying data by solving a pre-task [5]. The pseudo-labels of the pre-task are automatically generated based on data attributes and task definition (without any expensive manual annotation). Therefore, defining a pre-task in self-supervision is a key choice. Existing self-supervised methods can be broadly classified based on their pre-tasks into (a) generative methods, (b) context-based methods, and (c) cross-modal methods.

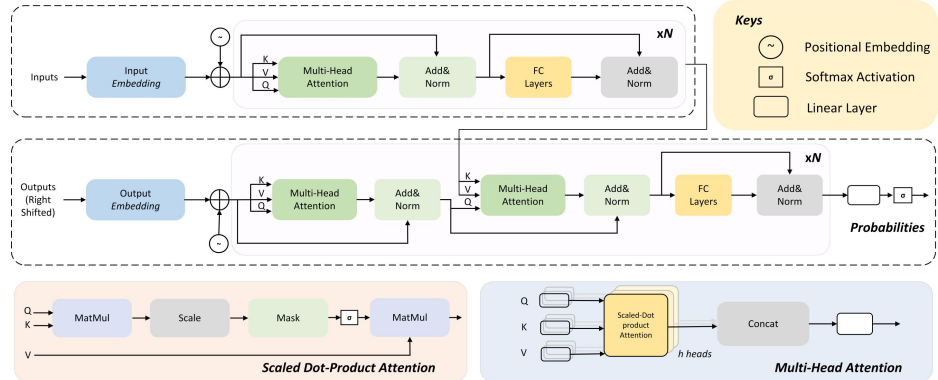

**Fig. 2.** The architecture of the Transformer model [1]. This model was originally developed for language translation tasks, where an input sequence in one language needs to be transformed into an output sequence in another language. The Transformer encoder (middle row) processes the input language sequence and converts it into embeddings before passing it on to the encoder modules. The Transformer decoder (bottom row) processes the previously generated outputs in the translation language and the encoded input sequences from the middle branch, to output the next word in the output sequence. The previously outputted sequence (used as the input to the decoder) is obtained by shifting the output sentence to the right by one position and adding a sentence-start token at the beginning. This shift prevents the model from simply learning to copy the decoder input to the output. The real data used to train the model is the output language sequence (without the right shift), with a sentence-end token appended. Blocks composed of multi-head attention (top row) and feed-forward layers are repeated N times in both the encoder and the decoder.

## 2.3   Transformer Model

The architecture of the Transformer model proposed in [1] is shown in Fig. 2. It has an encoder-decoder structure. The encoder (first row) consists of six identical modules (denoted as $N = 6$ in Fig. 2), each containing two sub-layers: a multi-head self-attention network, and a simple position-wise fully connected feed-forward network. After each module, residual connections [44] and layer normalization [45] are used.

It is noteworthy that, unlike conventional convolutional networks which perform feature aggregation and transformation simultaneously in the same step (e.g., through a convolutional layer followed by a non-linear operation), in the Transformer model, these two steps are decoupled, i.e., the self-attention layer only performs aggregation, while the feed-forward layer performs transformation. Similar to the encoder, the decoder in the Transformer model (bottom row) also contains six identical modules. Each decoder module has three sub-layers, the first two (multi-head self-attention and feedforward) are similar to those in the encoder, while the third sub-layer performs multi-head attention on the output of the encoder modules.

The original Transformer model [1] was trained for machine translation tasks. The input to the encoder is a sequence of words (i.e., a sentence) in one language.

Positional encoding is added to the input sequence to capture the relative positions of each word in the sequence. The dimension of the positional encoding is the same as the input, $d = 512$, and can be learned or predefined, e.g., through sine or cosine functions. As an autoregressive model, the Transformer model's [1] decoder uses previous predictions to output the next word in the sequence. Therefore, the decoder takes input from both the encoder and the previous outputs to predict the next word in the translated language sentence. To facilitate residual connections, the output dimension of all layers remains the same, i.e., $d = 512$. The dimensions of the query, key, and value weight matrices in multi-head attention are set to $d_q = 64, d_k = 64, d_v = 64$.

## 3    Datasets

Tactile datasets generally originate from various sensing technologies and can be categorized based on their characteristics. These datasets can typically be characterized by sensing modes, sensing resolution, tactile representation, and application scenarios. Sensing modes refer to the types of tactile information captured by the dataset, such as force, temperature, or texture. Sensing resolution describes the dataset's ability to distinguish tactile details. Tactile representation involves how the dataset expresses tactile information, for example, through pressure maps, deformation maps, or other sensor data. Application scenarios are specific tactile applications for which the dataset can be used, such as robotic haptics, virtual reality, etc. Here are some common tactile datasets:

**Multimodal Grasp Dataset** [46]**.** This dataset is designed for robotic manipulation research, featuring a combination of visual and tactile data. It encompasses 2,550 sets of valid data, derived from nearly 3,000 test trials using 10 different objects. The dataset provides a detailed account of both successful and failed grasping attempts, with a total success rate of 66.27%. It includes tactile data from 16-channel sensors across 400 steps in 24 seconds, along with visual data consisting of four images per grasp from two different camera viewpoints. Additionally, it captures the position sequence of eight motor joints and contains video data comprising two RGB and one depth video for each grasp. This rich dataset serves as a comprehensive tool for understanding the nuances of robotic grasping and manipulation, offering valuable insights into slip detection, and object classification.

**ObjectFolder series datasets**

- ObjectFolder 1.0 [47]: Introduced in 2021, this dataset includes 100 virtualized objects, each encoded with visual, auditory, and tactile sensory data. It's primarily focused on multisensory object recognition tasks and uses a uniform, object-centric, and implicit representation for each object's visual textures, acoustic simulations, and tactile readings. This makes the dataset flexible for use in various applications like instance recognition, cross-sensory retrieval, 3D reconstruction, and robotic grasping. However, it is noted for its limited scale and the quality of multisensory data, which might affect its generalization to real-world scenarios.

**Table 1.** Comparative Summary of Different Datasets Used in Tactile Domain Tasks. Each dataset is evaluated based on several parameters such as the number of object instances, total touches, data source, application in real-world scenarios, the environment in which data was collected, and the types of sensors used.

| Dataset | Object inst. | Touches | Source | Real-world | Environment | Sensor |
|---|---|---|---|---|---|---|
| Multimodal Grasp [46] | 3971 | 2550 | Robot | ✓ | Tabletop | Eagle Shoal Hand |
| ObjectFolder 1 [47] | 100 | N/A | Synthetic | ✗ | N/A | DIGIT [31] |
| ObjectFolder 2 [48] | 1000 | N/A | Synthetic | ✗ | N/A | GelSight [28] |
| ObjectFolder-Real [49] | 100 | N/A | Synthetic | ✓ | Indoor | GelSight [28] |
| Touch and Go [50] | 3971 | 13.9k | Human | ✓ | Indoor | GelSight [28] |

- ObjectFolder 2.0 [48]: Launched in 2022, this dataset expands significantly on its predecessor. It features 1,000 multisensory neural objects, offering a tenfold increase in the number of objects and substantial improvements in rendering time and multisensory data quality. These enhancements aim to address the limitations of OBJECTFOLDER 1.0, particularly regarding scale and data quality. ObjectFolder 2.0 is used to facilitate research in multisensory learning, particularly in computer vision and robotics, and supports tasks like object scale estimation, contact localization, and shape reconstruction.

- ObjectFolder-Real [49]: This dataset contains multisensory measurements for 100 real-world household objects. It builds upon a newly designed pipeline for collecting 3D meshes, videos, impact sounds, and tactile readings of these objects. The focus here is on providing a realistic and practical representation of objects, incorporating data collection processes tailored for each sensory modality. This dataset plays a crucial role in multisensory object-centric learning, particularly in tasks like object recognition, reconstruction, and manipulation.

**Touch and Go Dataset** [50]. This dataset is a rich collection of natural vision-and-touch signals, featuring approximately 13.9k detected touches and around 3971 individual object instances. Gathered by human collectors in diverse environments, it encompasses a broad spectrum of objects, both rigid and deformable, located in diverse settings that include indoor spaces like university buildings and apartments, as well as outdoor environments such as hiking trails and playgrounds. A significant feature of this dataset is its use of the GelSight tactile sensor, which records detailed tactile and visual information. The dataset's diversity, covering a broad range of real-world scenes and object types, makes it particularly valuable for research in multimodal learning. It's further enhanced by annotations for material categories and specific frame tagging, which are crucial for in-depth studies of visuo-tactile interactions in natural environments.

The section effectively summarizes various tactile datasets, each tailored for specific research applications in robotic manipulation and multimodal learning. These datasets, detailed in Table 1, range from the Multimodal Grasp Dataset with its focus on robotic manipulation, to the ObjectFolder series that progresses from virtualized objects to real-world scenarios, and the Touch and Go Dataset that captures natural vision-and-touch signals in diverse environments. Each dataset contributes unique insights and data, enriching the field of tactile sensing and offering valuable resources for research and development in this area.

## 4 Tasks and Transformers

### 4.1 Touch-Centric Tasks and Tactile-Only Transformers

Firstly, we review tasks and the corresponding Transformer-based methods that only use tactile data.

#### 4.1.1 Object Recognition

For the task of object surface recognition, Awan et al. [51] proposed a novel tactile texture classification method based on the Transformer architecture. In their study, the authors collected acceleration signals from 9 samples, which, after preprocessing and normalization, were segmented into 500 sample windows in the time domain as inputs to the model. After positional encoding, these were fed into three encoder layers, each with three attention heads. Notably, the model built by the authors only deployed the encoder module of the transformer, without using the decoder module. Each encoder layer is identical, containing two sub-layers: a self-attention sub-layer and a feedforward sub-layer. The feedforward sub-layer uses 1D-CNN and ReLU as the activation function, followed by a fully connected layer. Additionally, a normalization layer follows each sub-layer. For predictions, the output of the encoder module is fed into a Dense layer, followed by a softmax layer. Ultimately, the model proposed by the authors achieved an average accuracy of 98.87% and an F1 score of 96.89%, reaching state-of-the-art classification accuracy in texture classification tasks.

#### 4.1.2 Object Manipulation

For the object manipulation task in grasping, non-destructive grasping of deformable objects has always been a challenge in the field of robotics. Humans can perceive the physical properties of objects and apply precise forces while grasping fragile targets to accomplish dexterous and non-destructive operations. To transfer this capability from humans to robots, an improved Transformer model for processing complete tactile time series during the grasping process was proposed in [52]. Also, considering the scarcity of grasping tactile datasets available in the tactile domain compared to computer vision, the authors also established a tactile dataset of 9375 grasps of 15 types of fruits. The proposed model achieved a fruit recognition accuracy of 97.33% this dataset, surpassing traditional recurrent neural network (RNN) models. The results suggest that this work could facilitate tactile perception and non-destructive grasping in robots in agricultural fields, aiding in fruit picking, processing, sorting, and other related areas. This method also provides a technical reference for other studies on grasping tactile data.

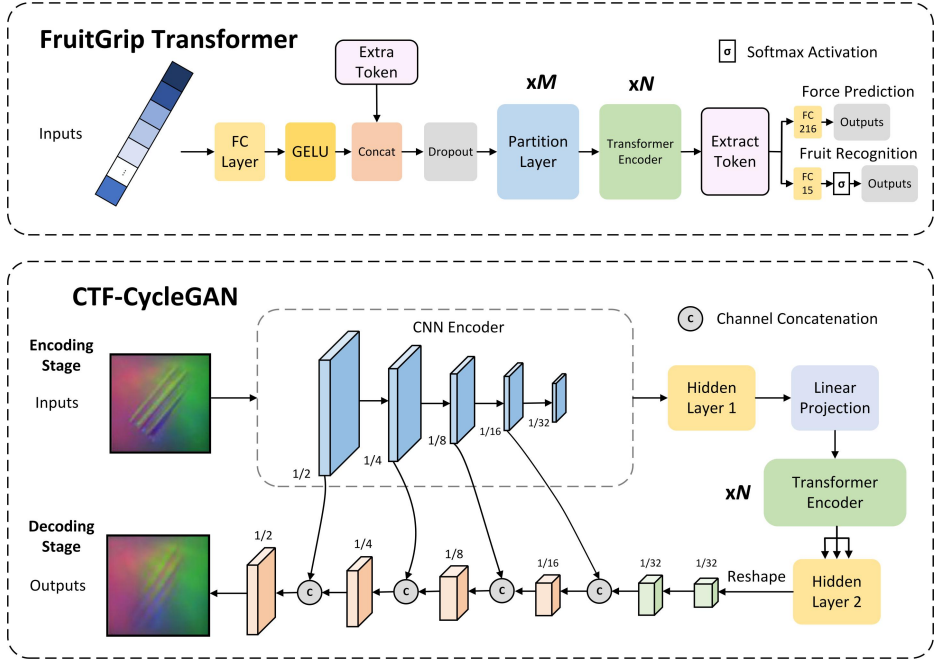

**Fig. 3.** Structural Display of the FruitGrip Transformer and CTF-CycleGAN Models for Object Manipulation Tasks.

For the task of manipulating tubular objects like test tubes, developing robots capable of handling them has the potential to accelerate experimental processes. However, for robots, learning to process tubular objects with a single perception and bridging the gap between simulation and reality remains a challenge. For this purpose, Zhao et al. [53] proposed a Sim2Real transferable hand posture estimation method based on tactile images, specifically for manipulating tubular objects. The core of this method is the CTF-CycleGAN model, a novel pixel-level unsupervised domain adaptation network that cascades CNNs and Transformers, aimed at narrowing the pixel-level domain differences in tactile tasks by introducing attention mechanisms and task-related constraints. Eventually, this method was implemented in a reinforcement learning-based policy learning framework for learning Sim2Real transferable robot insertion and removal actions, further applied in human-robot collaborative test tube placement scenarios and robotic pipetting scenes. Experimental results showed that the learned tactile-motor strategy has generalization capabilities for tubular object manipulation.

To better illustrate the FruitGrip Transformer and CTF-CycleGAN models, Fig. 3 depicts their structural designs. As shown, the FruitGrip Transformer model uses a framework similar to the Visual Transformer (ViT) but with specific adaptations for tactile data. The model directly projects one-dimensional tactile sequences of 216-dimensional vectors to a specified dimension and then processes them using a Transformer encoder. Uniquely, it segments the input sequence regionally, simulating tactile perception in different parts of a human hand, and then processes it through Transformer blocks of various scales. This structure helps the model effectively learn and predict the type and grasping force of fruits, facilitating dexterous manipulation of deformable objects.

In contrast, the CTF-CycleGAN model combines Convolutional Neural Networks (CNNs) and Transformers in an encoder-decoder structure. In the encoding phase, it first uses multiple convolutional layers to extract high-level features from tactile images, then enhances feature extraction capabilities with a self-attention Transformer encoder. Its distinguishing feature is the ability to capture global information in tactile images, focusing on regions with significant features. The decoding phase uses a U-shaped structure to combine features from the Transformer encoder and CNNs for image generation.

In summary, the FruitGrip Transformer is primarily focused on precise object recognition and force prediction through the analysis of tactile time series data, while the CTF-CycleGAN is dedicated to the generation of high-quality images and advanced feature extraction, with a specific emphasis on tactile imagery processing. Although both models incorporate Transformer technology, they are distinct in their application objectives and structural designs, showcasing the versatility and adaptability of deep learning in diverse robotic applications.

### 4.1.3 Others

Compared to other multimedia applications like videos, sounds, and images, the transmission of tactile signals requires higher reliability and lower latency. Vibrational tactile signals can represent key elements of rich tactile information, such as textured surfaces. To transmit these signals to remote locations, and create immersive and realistic user experiences, the development of an efficient tactile compression method is needed. In [54], a Transformer-based vibrational tactile signal codec TVSC was proposed, combining Transformer structure, quantization, and entropy coding. It is trained end-to-end for the encoder and decoder to achieve efficient compression of tactile signals. After inverse quantization, the decoder is fine-tuned to overcome the impact of signal loss caused by quantization. When the compression ratio is small, the signals reconstructed by TVSC closely match the original signals. As the compression ratio increases, the quality of the reconstructed signals slightly decreases and stabilizes. Experimental results show that TVSC outperforms PVC-SLP [55].

**4.2    Visuo-Tactile Tasks and Visuo-Tactile Transformers**

Tasks that combine sight and touch involve the combination of visual and tactile sensory inputs to carry out different functions or accomplish specific goals. These tasks led to the development of Visuo-Tactile Transformers.

We review tasks and the corresponding Transformer-based methods that integrate visual and tactile sensory modes.

**4.2.1 Object Recognition**

In the field of computer vision, there has been extensive research and application of object recognition technology. Specifically, 2D and 3D images have been effectively used for classifying and recognizing objects and their surface properties. For instance, convolutional neural networks (CNNs) have been successfully applied to signal recognition processes [56]. Moreover, the application of one-dimensional CNNs and bidirectional long short-term memory networks (Bi-LSTMs) in-vehicle network anomaly detection was demonstrated in [57].

Unlike touch-based methods, vision-based methods require the use of optical sensors, which may be limited by obstructions and lighting conditions [58]. Conversely, methods that combine both visual and tactile modalities have been proven to outperform methods using only the visual modality [59].

However, integrating visual and tactile features faces significant challenges. Since visual and tactile features exist in different feature spaces, it can be challenging for fusion modules to establish direct interactions between them. Current methods handle visual and tactile features during the fusion stage but struggle to bridge the significant gap between these two modalities. For instance, visual data and tactile data vary in terms of frequency, signal type, and perceptual domain.More specifically, visual sensors like cameras collect data through low-frequency instantaneous captures, while tactile sensors acquire data through high-frequency continuous sliding or touching. The signal type of visual data consists of RGB images composed of pixels, while tactile data includes various signal types, such as vibration signals, force feedback, and tactile images. Regarding the perceptual domain, visual perception focuses on the perception of shape and local color, while tactile perception emphasizes pressure and roughness. All these differences make it challenging to directly fuse information from these two modalities.

To address the aforementioned issues, Wei et al. [60] proposed a new visual-tactile Transformer-based method based on Alignment and Multi-Scale Fusion (AMSF). First, a visual-tactile contrastive learning module is used to match visual and tactile data, and then the matched features are fed into a multi-scale fusion module through a Transformer, achieving deep interaction between the two modalities. By leveraging these two modules, this method not only aligns visual and tactile modality but also achieves a fusion of multi-scale feature maps. Extensive experiments on three public datasets (GelFabric [28], LMT [61, 62], and Vitac [22]) demonstrated the superiority of the proposed method. Subsequently, Li et al. [63], to address the challenge of fusing visually and tactilely information, designed a visual and tactile information fusion mechanism based on the Transformer architecture, named VITO-Transformer. Thanks to this fusion mechanism, the accuracy of object recognition was significantly improved. In the end, comprehensive comparative experiments were conducted on the public visual-tactile dataset (ObjectFolder 2.0) and a custom visual-tactile dataset (700 object visual images and 7 object tactile images), confirming the advantages of VITO-Transformer and validating the effectiveness of this fusion mechanism compared to current popular fusion algorithms, bringing innovative solutions to the field of visual-tactile fusion.

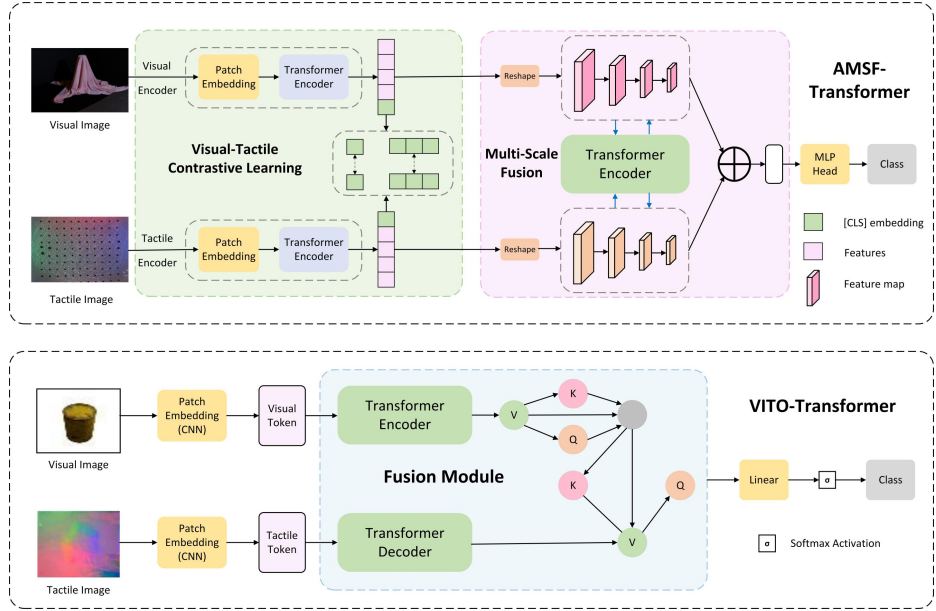

**Fig. 4.** Structural Display of the AMSF-Transformer and VITO-Transformer Models in Object Manipulation Tasks.

To delve deeper into and compare the AMSF-Transformer with the VITO-Transformer, Fig. 4 presents their structural details. AMSF utilizes a contrastive learning approach to align visual and tactile information, followed by multi-scale fusion using a Transformer. The crux of this model lies in using a visual Transformer as a dual-modal encoder to process input visual and tactile images. The alignment of these two representations is achieved through visual-tactile contrastive loss. Its multi-scale fusion module operates at every layer of the multi-modal encoder, ensuring effective feature fusion, and ultimately processes the fused features through an MLP for classification.

In contrast, The VITO-Transformer model adopts an innovative decoding mechanism for the fusion of visual and tactile data. The model initially segments and transforms the input visual image into token embedding sequences, which are then fed into the Transformer encoder. Additionally, tactile data is converted into tactile image data and similarly segmented and transformed into token embedding sequences. These sequences, processed by the Transformer encoder, are subsequently input into the decoder. The VITO-Transformer's decoder crucially integrates visual and tactile information by adding an extra layer of multi-head attention operations. This unique mechanism allows the decoder to combine and process these two different types of information effectively, generating fused feature information for precise identification of various objects.

Although both models are dedicated to the fusion of visual and tactile information, they differ in their implementation methods and structures. AMSF focuses on information alignment and layered fusion, while VITO-Transformer emphasizes information fusion through its specialized decoder mechanism. This comparison reveals the diverse approaches to multi-modal perception in the field of machine learning, providing valuable insights for future research.

In the field of autonomous driving, to achieve accurate road recognition (object recognition task for identifying specific roads), a multimodal fusion recognition network based on the CNN-Transformer structure was proposed in [64]. Specifically, visual and tactile channels were input into specific channel SE-CNNs, which emphasized valuable input information to obtain weighted features. These features were then input into a ``bottleneck'' fusion Transformer encoder and produced recognition results. Additionally, the authors designed a fusion feature extractor to enhance fusion representation capabilities and improve fusion accuracy. To test the model's performance, the authors conducted vehicular field experiments and established a dataset composed of four types of road surfaces, demonstrating that the network achieved an accuracy of 99.48% in road recognition tasks, showing its strong application value.

**4.2.2 Cross-Modal Generation**

For cross-modal generation tasks, fully capturing the underlying correlations between multimodal data and generating more accurate tactile data for high-fidelity tactile rendering remains a challenging issue. To address this, a vision-audio assisted tactile Transformer model for cross-modal generation of tactile friction coefficients was proposed in [65]. The model first encodes the time-frequency graph of audio and the RGB image input of vision to extract joint encoding features. Then, these joint encoding features are transformed into tactile decoding features by a Transformer module.After that, the tactile decoding features are used to reconstruct tactile amplitude and phase spectra. Finally, they are converted into 1-D friction coefficients through Inverse Short Time Fourier Transform (ISTFT) [66]. This method combines self-attention mechanisms and convolution operations to more comprehensively capture the correlations between audio-visual-tactile data. Ultimately, the model was evaluated on the LMT tactile materials dataset [56]. Results showed that the proposed tactile Transformer model outperformed AVHR [67] and convolutional autoencoder (CAE) methods [68], improving Root Mean Square Error (RMSE) by 28.53% and 33.25%, respectively. Comparative experimental results also indicated that lighting conditions affect the accuracy of the vision-tactile model, with RMSE for low-light images being 24.10% higher than for normal images, but the combination of audio and low-light images could compensate for errors caused by lighting conditions. Additionally, the tactile Transformer model proposed by the authors achieved a tactile rendering fidelity of 92.3% compared to AVHR and CAE methods, significantly enhancing rendering fidelity ($p < 0.05$). This result was obtained through a specific statistical significance test, namely the Analysis of Variance (ANOVA). However, the ability of the model to generate materials for new categories could not be fully verified due to the unbalanced distribution of different material categories in the existing datasets.

For specific domain cross-modal generation, models combining visual and tactile modalities based on the Transformer architecture also play an important role. For example, in the VR/AR field, a method for generating tactile signals based on the Transformer network framework was proposed in [69]. This method uses the visual image of an object's surface as visual data and scanning parameters (scanning speed $v$, position $p$, and normal force $f$) generated by a pen sliding over the object's surface as acceleration signals. The acceleration signals are synthesized through a Transformer-based generation model and a multimodal feature embedding module as the tactile data for the object's surface. Unlike traditional Transformer structures applied in NLP, this model replaces the input/output embedding layers with linear layers to adapt to the size of the encoder/decoder inputs and uses a linear projection layer to reshape the final output of the decoder to the target size. Especially, this model also adds self-attention layers, multimodal embedding layers, and feedforward layers in the latent space to adapt to the multimodal fusion of visual-tactile data. To evaluate the model's performance, the authors used the aluminum texture as a test sample and employed a sliding window-based data augmentation strategy to create a multimodal dataset. Finally, the Root Mean Square Error (RMSE) value for the selected samples was 0.0082 (SD = 0.0023). In the future, the authors will conduct user studies to assess the realism of the generated virtual tactile textures and further integrate the tactile texture modeling and rendering framework with VR/AR devices to improve the user experience of texture simulation in virtual environments.

### 4.2.3 Object Manipulation

The object manipulation task is subdivided into four sub-tasks: 1) slip detection, 2) object grasping, 3) object rotation, and 4) combined manipulation of objects. Each sub-task's visuo-tactile Transformer architecture models will be introduced in detail.

In object manipulation tasks, slip detection plays a crucial role in dexterous robot gripping and manipulation and has always been a challenging problem in the field of robotics. However, due to limitations in perception capabilities, the desired detection performance may not be achievable through tactile perception alone. Adding visual perception is an intuitive solution. Li et al. [70] proposed a vision-tactile learning method based on deep neural networks to detect slip, highlighting the importance of vision-tactile fusion perception in slip detection tasks. Compact and multimodal representations of sensory inputs learned through self-supervised learning can be used to improve the sample efficiency of policy learning [71]. However, these early fusion methods, which directly combine features of the visual and tactile modalities, have been shown to lack the ability to capture modality-specific and cross-modal features [72]. Furthermore, most of the methods mentioned above are explicitly designed for tactile sensors with different data formats and sampling rates, leading to significant limitations in platform generalizability.

To address the inconsistency of vision-tactile sequence learning in slip detection tasks, Cui et al. [73] proposed a generalized vision-tactile Transformer (GVT-Transformer) model. Specifically, the GVT-Transformer mainly consists of modality-specific and cross-modality Transformers, capturing features of specific modalities and cross-modality, respectively. Experimental results showed that this method not only outperformed traditional early fusion methods in detection performance but also was more suitable for non-aligned situations. Slip detection experiments conducted on real robots also confirmed the Sim2Real generalization capability of the proposed method.

In the object grasping task, safe grasping of deformable objects remains a challenging task. Loose grasps with small grasping forces can lead to object slip, while large grasping forces can cause object damage. Additionally, the contact geometry shape and friction characteristics of an object might affect the optimal grasping force for safe grasping. To address this issue, a generalizable vision-tactile robotic grasping method based on the Transformer architecture was proposed in [37]. In this method, the Transformer model performs two predefined exploration actions (pinching and sliding), not only predicts the outcome of various grasping intensities using a multilayer perceptron (MLP) but also enhances its grasp prediction capabilities by learning physical feature embeddings derived from sensor feedback. Using these predictions, the gripper can infer safe grasping intensities. To evaluate the model's performance, the authors first conducted benchmark tests for slip detection tasks on a public dataset containing 84 everyday objects. The results showed that the model far surpassed the CNN+LSTM model in terms of grasping accuracy and computational efficiency. Besides, the authors collected a new dataset containing six types of fruits (i.e., plums, oranges, lemons, tomatoes, apples, and kiwifruits) for grasping, and used the model for online grasping experiments on both seen and unseen fruits. Finally, the authors extended the model to different-shaped objects and demonstrated the effectiveness of the pre-trained model trained on a large-scale fruit dataset.

In the object rotation task, accurate control of robot finger movements is required, and tactile feedback must be received and processed in real time to adjust rotation strategies. For this purpose, a vision-tactile Transformer-based model RotateIt was proposed in [74]. This model effectively processes multimodal perceptual information by integrating visual information provided by depth cameras and tactile data from tactile sensors. The authors first built a dataset of objects with different physical properties and random initial postures from EGAD, Google Scanned Objects, YCB, and ContactDB, with a width/depth/height ratio of less than 2.0. Visual information is processed by a convolutional neural network (ConvNet), while tactile information is transformed into a discrete contact point position representation. The RotateIt model combines these visual and tactile features to achieve multi-axial object rotation based on fingertip. Experimental results showed that RotateIt could effectively understand the geometric shape and contact point characteristics of objects, enabling robots to manipulate objects more naturally and effectively.

In the task of combined manipulation of objects, existing vision-tactile methods typically use latent representations that are vectors or clusters in $R^n$ and do not utilize the attention architecture [1]. These representations might flatten details and face difficulties in the locality of the images. These problems are particularly tricky for manipulation, as the difference between contact and non-contact might be small in pixel space, or the robot might need significant movement to make contact with an object. To address this issue, Chen et al. [75] proposed a model for object manipulation, the Visuo-Tactile Transformer, abbreviated as VTT. This model uses self-attention mechanisms and cross-modal attention mechanisms to fuse visual and tactile information and improves feature learning through learned embedding vectors. The authors also proposed three types of learnable embeddings to enhance model performance. Finally, the model was combined with reinforcement learning to form a stable and robust manipulation method. The model was experimentally evaluated on four simulated robotic tasks (Pushing, Door-Open, Picking, Peg-Insertion) and a real-world pushing task. Compared to baseline methods such as concatenation and product of expert (PoE), VTT demonstrated effectiveness in representation learning.

**4.2.4 Others**

In terms of multimodal representation, Noguchi et al. [76] proposed a model that learns visual, tactile, and proprioceptive perception to obtain multimodal plastic body and near-personal space representations. This approach includes three main stages: 1) encoding of visual, tactile, and proprioceptive perceptions, 2) integration of encoded perceptions through Transformer architecture utilizing a self-attention mechanism, 3) reconstruction of visual and tactile perceptions from integrated sensory representations. The model, focused on a simulated robot arm, facilitates the development of multimodal and body-centered peripersonal space representation through tool use, surpassing previous models that offered non-body-centered plastic body representations. By learning from camera vision, arm touch, and proprioception of camera and arm postures, it achieves a body representation that localizes tactile sensations on a concurrently developed peripersonal space representation. This learning process, especially during tool use, endows the body representation with plasticity, allowing it to adapt to changes in interaction brought by tool use while maintaining an internal spatial coordinate system. The research advances the understanding of multimodal representation development and the plasticity of self-body cognition, providing insights into cognitive neuroscience and developmental robotics.

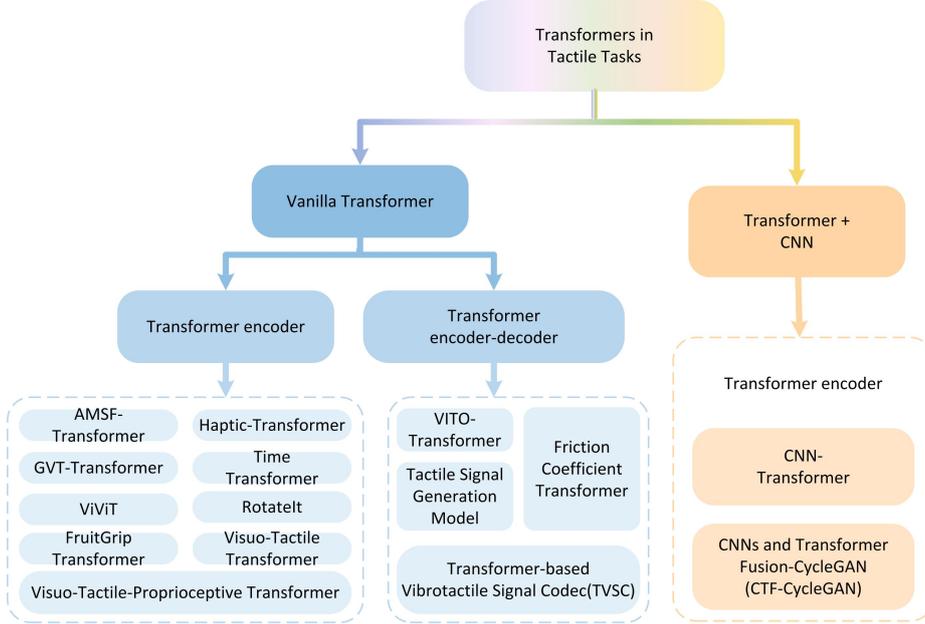

**Fig. 5.** Taxonomy of Transformer-based Approaches for Tactile Tasks: First divided into Vanilla Transformer and Transformer + CNN, then further subdivided into Transformer encoder and Transformer encoder-decoder.

## 5      Open Challenges and Future Directions

Despite the outstanding performance and intriguing features of Transformer models (as discussed in Table 2 and Fig. 5), applying them in practical settings still presents several challenges. The most significant bottlenecks include the need for large amounts of training data and the associated high computational costs. There are also challenges in visualizing and interpreting Transformer models. In this section, we outline these challenges, mention some recent efforts made to address these limitations and highlight open research questions.

### 5.1     High Computational Cost

As discussed in Sec.1, a key advantage of Transformer models is their flexibility to scale up to high parameter complexity. While this is a notable feature, allowing for the training of enormous models, it leads to higher training and inference costs. For instance, the training time for the base BERT model [5] (with 110 million parameters) is approximately 1.89 petaflop days, while the GPT-3 model [4] (with 175 billion parameters) requires an astonishing ~1925 times increase, approximately 3640 petaflop days. This comes with a hefty cost tag; for example, according to an estimate [77], GPT-3's training could have cost OpenAI 4.6 million. Furthermore, these large-scale models require significant compression (such as distillation) to be feasible in real-world settings.

**Table 2.** Comprehensive Overview of Transformer-based Approaches for Tactile Tasks, Including Method Names, Associated Tasks, Utilized Datasets, Evaluation Metrics, and Key Design Highlights.

| Method | Task | Datasets | Metrics | Design Highlights |
|---|---|---|---|---|
| AMSF-Transformer [38] | Object Recognition | GelFabric, LMT, Vitac | Accuracy, Precision | New alignment strategy, multimodal contrastive learning. |
| VITO-Transformer [63] | Object Recognition | OBJECTFOLDER 2.0, Self-made Dataset | Accuracy, Precision, Recall and F1 | Visual-tactile fusion via Transformer network. |
| CNN-Transformer [64] | Object Recognition | Four different kinds of roads Dataset | Accuracy | Multimodal fusion with CNN-Transformer, channel-specific feature weighting. |
| Haptic-Transformer [51] | Object Recognition | Fabric, steel, plastic and wood Dataset | Accuracy, Precision, Recall and F1 | Transformer model for tactile texture classification. |
| Friction Coefficient Transformer [65] | Cross-Modal Generation | LMT Haptic Material Dataset | RMSE, Relative Error | Self-attention for audio-visual encoding, tactile friction generation. |
| Tactile Signal Generation Model [69] | Cross-Modal Generation | Enhanced HaTT Dataset | RMSE | Visual-tactile input to simulate tactile textures in AR/VR. |
| Generalized Visual-Tactile (GVT) Transformer [73] | Object Manipulation | Visual-Tactile Dataset | Precision, Recall and F1 | Modality-specific and cross-modality Transformers for non-aligned situations. |
| TimeTransformer and ViViT [37] | Object Manipulation | Slip Detection: 84 daily objects. Safe Fruit Grasping: Involving six different types of fruits. | Accuracy, Success Rate | Self-attention for deformable object tasks, outperforms CNN+LSTM. |
| RotateIt [74] | Object Manipulation | EGAD, Google Objects, YCB, ContactDB | TTF, RotR, RotP and Rotations | Vision-tactile integration, depth cameras and tactile sensors fusion. |
| Visuo-Tactile Transformer (VTT) [75] | Object Manipulation | RGB images, wrist sensor data | Success Rate | Self and cross-modal attention, learnable embeddings for feature enhancement. |
| FruitGrip Transformer [52] | Object Manipulation | Fruit grasping dataset | Accuracy, Speed and Parameters | Processes tactile time series, tactile dataset establishment. |
| CTF-CycleGAN [53] | Object Manipulation | Real-world tactile images Dataset | Success Rate | Pixel-level domain adaptation, CNN-Transformer cascade, attention mechanisms. |
| Visuo-Tactile-Proprioceptive Transformer [76] | Others | Visuo-tactile-proprioceptive Dataset | PC space visualization | Body representation model with tool-use-induced plasticity. |
| Transformer-based Vibrotactile Signal Codec (TVSC) [54] | Others | TUM Dataset and LMT tactile material Dataset | PSNR, Compression Ratio | Vibrotactile signal compression using Transformer-based codec. |

### 5.2   Large Data Requirement

Because Transformer architectures do not inherently encode inductive biases (i.e., prior knowledge) for processing tactile data, they typically require extensive training to figure out the modality-specific underlying rules. For instance, CNNs have built-in translation invariance, weight sharing, and local scale invariance due to pooling operations or multi-scale processing blocks. However, Transformer networks need to figure out these tactile data-specific concepts from training examples on their own, leading to longer training times, a significant increase in computational demands, and the need for large datasets to process.

### 5.3   Touch-Tailored Transformer Design

It is observed that the majority of previous studies in the field of tactile tasks tend to directly utilize NLP Transformer models for addressing problems in the tactile domain. These include architectures designed for object recognition, object manipulation and especially multi-modal processing. While the preliminary results of these straightforward applications are quite encouraging and motivate further exploration of the advantages of self-attention and self-supervised learning, the current architectures may still be more suited to language problems (using sequential structures) and require further intuition to make them more effectively applicable to tactile inputs.

### 5.4   Hardware Efficient Design

Large Transformer networks can have dense power and computational demands, hindering their deployment on edge devices and in resource-constrained environments (such as internet-of-things platforms).

This challenge becomes even more critical in the tactile field, where the need for real-time haptic feedback imposes stringent requirements on latency and processing speed. The Tactile Transformer, which facilitates the transmission of touch and manipulation in real-time over digital networks, demands not only low latency but also high reliability and rapid processing to simulate real-world physical interactions accurately.

Recent reports suggest efforts in compressing and accelerating NLP models on embedded systems like Field-Programmable Gate Arrays (FPGAs) [78]. Li et al. [78] used an enhanced block circulant matrix-based representation to compress NLP models and proposed a new FPGA architecture design to efficiently manage resources for high throughput and low latency. They were able to achieve 27-, 3-, and 81-times performance improvement (measured in FPS throughput) over the RoBERTa model [41] on CPU, reducing power consumption and energy efficiency.

To achieve this goal, Wang et al. [79] proposed using a neural architecture search strategy to design Hardware Aware Transformers (HAT) [80, 81, 82] Specifically, a SuperTransformer model [79] is first trained as a performance surrogate, capable of estimating model performance without full training. This model contains the largest possible model in the search space and shares weights across common parts. Ultimately, an evolutionary search is performed, considering hardware latency constraints, to find suitable SubTransformer models [79] for the target hardware platform (e.g., internet-of-things (IoT) devices, GPU, CPU). However, such hardware-efficient designs are currently lacking, to enable seamless deployment of Tactile Transformers on resource-limited devices. Moreover, the search cost of evolutionary algorithms remains substantial due to the associated environmental impact of $CO_2$ emissions.

## 5.5 Towards Integrating All Modalities

Given that transformers provide a unified design to process different modalities, recent efforts have also focused on proposing more general-purpose, transformer-based universal reasoning systems. Inspired by biological systems capable of processing different forms of information, the Perceiver model [83] aims to learn a unified model that can handle any given input modality without making specific domain structural assumptions. The perceiver employs an asymmetric cross-attention approach to distill input information into low-dimensional latent bottleneck features to scale to high-dimensional inputs. Once features are refined into a compact and fixed-dimensional form, regular Transformer blocks are applied in the latent space. The initial Perceiver model demonstrated performance competitive with ResNets [84] and ViTs [12] in image classification and could handle 3D data, audio, images, videos, or their combinations. However, the model could only generate fixed outputs like class probabilities.

A recent improvement, termed Perceiver IO [85], aims to learn models with flexible inputs and arbitrary-sized outputs. This allows applications to problems requiring structured outputs, like natural language tasks and visual understanding. While these models avoid reliance on modality-specific architectural choices, the learning itself still involves modality-dependent choices, such as specific augmentations or positional encodings. An interesting and open direction for the future is to achieve complete modality agnosticism.

## 6      Conclusion

This review begins with a detailed introduction to the two fundamental ideas of the Transformer model and its traditional architecture. It then delves into the application of the Transformer architecture in various tactile tasks, such as object recognition, cross-modal generation, and object manipulation. The approaches discussed mainly fall into two categories: models that rely solely on tactile data and those that integrate tactile and visual data.

Our exploration encompassed not only the architectural innovations and their applications but also the challenges and potential future developments. Key challenges such as the high computational cost, large data requirements, and the need for touch-tailored Transformer designs were discussed, providing a comprehensive overview of the current state and prospects of Transformer models in the tactile domain.

We hope that this effort will further stimulate interest in the tactile community to harness the potential of Transformer models and to improve their current limitations.

**Acknowledgments.** This work is supported by the Talent Fund of Beijing Jiaotong University (2023XKRC006) and the Pattern Recognition Center, WeChat AI, Tencent Inc.

**Disclosure of Interests.** The authors have no competing interests to declare that are relevant to the content of this article.